%% file: submission.tex
\documentclass[11pt]{article}  

\newcommand{\xv}{\boldsymbol{x}}
\newcommand{\xnew}{\xv_{\mathrm{new}}}
\newcommand{\yv}{\boldsymbol{y}}
\newcommand{\yall}{\yv_{1:N}}
\newcommand{\fv}{\boldsymbol{f}}
\newcommand{\fall}{\fv_{1:N}}
\newcommand{\zerov}{\boldsymbol{0}}
\newcommand{\thetav}{\boldsymbol{\theta}}
\newcommand{\Lcomp}{L_{\mathrm{comp}}}
\newcommand{\appxref}[1]{Appendix~\ref{#1}}
\newcommand{\thmref}[1]{Theorem~\ref{#1}}
\newcommand{\trace}{\mathrm{tr}}
\newcommand{\vectorize}{\mathrm{vec}}
\newcommand{\pd}[2]{\frac{\partial #1}{\partial #2}}

\input{settings/preamble.tex}

\begin{document}
\input{settings/author-info.tex}
\input{sections/abstract/main.tex}

\section{Introduction}

Gaussian process (GP) is one of the most important machine learning algorithms in practice and often plays a key role in Bayesian optimization (BO)~\citep{brochu2010tutorial,shahriari2016taking,garnett2022bayesian}~\footnote{
  BO is not necessarily dependent on GP.
  For example, TPE~\citep{bergstra2011algorithms,watanabe2023tree} uses kernel density estimation and SMAC~\citep{lindauer2022smac3} uses random forest.
} because GP shows good predictive accuracy even with a small amount of data.
While the vanilla GP models only one output, multi-output modeling by GP~\citep{bonilla2007multi}, aka multi-task GP (MTGP), often brings benefits as represented by \cite{swersky2013multi,daulton2020differentiable,daulton2022multi}.
Although they demonstrated that MTGP is effective for multi-objective optimization, constrained optimization, multi-fidelity optimization, and meta-learning, many works still do not rely on MTGP, e.g., multi-objective optimization by \cite{yang2019multi}, constrained optimization by \cite{gardner2014bayesian,gelbart2014bayesian,eriksson2021scalable}, multi-fidelity optimization by \cite{kandasamy2017multi,song2019general,kandasamy2019multi,wistuba2022supervising}, and meta-learning by \cite{feurer2018practical}.
That is partially because the original paper~\citep{bonilla2007multi} unfortunately lacks their derivation details, making it challenging to fully understand.
To this end, we remove this barrier in this paper by giving more detailed and friendly derivations of the formulations and the gradients of the formulations with respect to kernel hyperparameters.
By doing so, we would like more researchers to work on the aforementioned problem setups using MTGP, potentially leading to further enhancements in existing hyperparameter optimization frameworks that use MTGP such as BoTorch~\citep{balandat2020botorch}.

\section{Related Work}
In this paper, we focus only on MTGP formulated in \cite{bonilla2007multi} simply because this model is used in BoTorch~\footnote{
  See the documentation string of \texttt{MultiTaskGP} in \url{https://botorch.org/api/_modules/botorch/models/multitask.html}.
}, which is the most widely used GP-based BO framework we are aware of.
\cite{bonilla2007multi} formulated MTGP as a linear combination of latent GPs with fixed coefficients as explained later.
This formulation is relatively simple and a large body of existing work tackled to capture more complex structures.
For example, \cite{wilson2011gaussian} modeled the coefficients dependent on input variables, \cite{alvarez2008sparse} employed convolved process, and \cite{titsias2009variational,nguyen2014collaborative} used variational inference to approximate the posterior of the latent processes.

\section{Notations}
In this paper, we use the following notations:
\begin{enumerate}
  \item $\mathcal{N}(\boldsymbol{\mu}, \Sigma)$, the Gaussian distribution with the mean $\boldsymbol{\mu}$ and the covariance matrix $\Sigma$,
  \item $\mathcal{N}(\boldsymbol{z} | \boldsymbol{\mu}, \Sigma)$, the probability density function of the Gaussian distribution with the mean $\boldsymbol{\mu}$ and the covariance matrix $\Sigma$,
  \item $\xv \in \mathcal{X} \subseteq \mathbb{R}^D$, an input vector $\xv$ defined on a $D$-dimensional domain $\mathcal{X}$,
  \item $f_m: \mathcal{X} \rightarrow \mathbb{R}$, the unobservable output mean function of the $m$-th output,
  \item $M \in \mathbb{Z}_+$, the number of outputs given an input vector,
  \item $y_{n,m} \sim \mathcal{N}(f_m(\xv_n), \sigma_m^2)$, the observed $m$-th output value of the $n$-th input vector $\xv_n$,
  \item $\Sigma \coloneqq \mathrm{diag}[\sigma_1^2, \dots, \sigma_M^2] \in \mathbb{R}^{M \times M}$, a diagonal matrix with the $(m, m)$-th element $\sigma_m^2$,
  \item $\zerov_N \in \mathbb{R}^{N}$, a zero vector with the size $N$,
  \item $I_N \in \mathbb{R}^{N \times N}$, an identity matrix with the shape of $N \times N$,
  \item $S \coloneqq I_N \otimes \Sigma \in \mathbb{R}^{NM\times NM}$, the Kronecker product of $I_N$ and $\Sigma$,
  \item $\yv_n \coloneqq [y_{n, 1}, y_{n, 2}, \dots, y_{n,m}] \in \mathbb{R}^{M}$, the observed output vector given the $n$-th input vector,
  \item $Y \in \mathbb{R}^{N \times M}$, the observed output matrix with the $(i, j)$-th element $y_{i,j}$,
  \item $\fv_n \coloneqq [f_{n,1}, \dots, f_{n,m}] \coloneqq [f_1(\xv_n), f_2(\xv_n), \dots, f_m(\xv_n)] \in \mathbb{R}^{M}$, the output mean vector given the $n$-th input vector,
  \item $F \in \mathbb{R}^{N \times M}$, the output mean matrix with the $(i,j)$-th element $f_{i,j}$,
  \item $\yall \coloneqq \vectorize(Y^\top) = [\yv_1^\top, \dots, \yv_N^\top]^\top \in \mathbb{R}^{NM}$, the flattened observed output vector,
  \item $\fall \coloneqq \vectorize(F^\top) = [\fv_1^\top, \dots, \fv_N^\top]^\top \in \mathbb{R}^{NM}$, the flattened output mean vector,
  \item $k_{\thetav}: \mathcal{X} \times \mathcal{X} \rightarrow \mathbb{R}$, a kernel function given its hyperparameters $\thetav$,
  \item $K_x \in \mathbb{R}^{N \times N}$, a kernel matrix for the input vectors with the $(i, j)$-th element $k_{\thetav}(\xv_i, \xv_j)$,
  \item $K_f \in \mathbb{R}^{M \times M}$, a kernel matrix for outputs,
  \item $K_{xf} \coloneqq K_x \otimes K_f \in \mathbb{R}^{NM \times NM}$, the Kronecker product of $K_x$ and $K_f$, and
  \item $A_{i,j} \in \mathbb{R}$, the $(i,j)$-th element of the matrix $A$.
\end{enumerate}
Note that we assume that kernel matrices are positive definite and symmetric, meaning that they are invertible, and the output function $f_{\cdot}: \{1,\dots, M\} \times \mathcal{X} \rightarrow \mathbb{R}$ follows the Gaussian process, i.e., $f_{\cdot} \sim \mathcal{GP}(\mu, k_{\thetav}k_f)$ where $k_f(i,j) = (K_f)_{i,j}$ is an index kernel for the output correlation, and $\vectorize: \mathbb{R}^{N \times M} \rightarrow \mathbb{R}^{NM}$ is the vectorization operation:
\begin{equation}
  \vectorize(C^\top) = \vectorize\begin{pmatrix}
  \begin{bmatrix}
    C_{1,1} & \cdots & C_{1,M} \\
    \vdots & \ddots & \vdots \\
    C_{N,1} & \cdots & C_{N,M} \\
  \end{bmatrix}
  \end{pmatrix}
  = [C_{1,1}, \dots, C_{1,M}, C_{2,1}, \dots, C_{2,M}, \dots, C_{N,1}, \dots, C_{N,M}]^\top.
\end{equation}
Notice that $\vectorize(C^\top)$ is equivalent to \texttt{C.flatten()} in the NumPy style.
Furthermore, we define the Kronecker product $\otimes$ as follows:
\begin{equation}
\begin{aligned}
  A \otimes B = \begin{bmatrix}
      A_{1,1}B & \cdots & A_{1,N}B \\
      \vdots & \ddots & \vdots \\
      A_{N,1}B & \cdots & A_{N,N}B \\
    \end{bmatrix} \in \mathbb{R}^{NM \times NM}
\end{aligned}
\end{equation}
where $A \in \mathbb{R}^{N \times N}$ and $B \in \mathbb{R}^{M \times M}$.

\section{Output Correlation Inference for Multi-Task Gaussian Process}

\cite{bonilla2007multi} proposed to model the interaction effects between each output by assuming $\mathcal{N}(\fall | \zerov_{NM}, K_{xf})$ as the prior and $\mathcal{N}(\yv | \fv, \Sigma)$ as the likelihood.
Importantly, we need to estimate the kernel matrix $K_f$ for outputs and \cite{bonilla2007multi} introduced the EM algorithm~\citep{mclachlan2008algorithm} update for this and the gradient approach.
However, the paper unfortunately lacks their derivation details.
This section provides the derivation details to fill the gap in the original paper.
Note that the approach used in BoTorch is the gradient approach explained in \secref{main:method:gradient-approach}.

\subsection{EM Algorithm Update for Multi-Task Gaussian Process}

We first provide a friendly derivation of the EM algorithm update for MTGP.

\subsubsection{Complete-Data Log-Likelihood}

Since MTGP requires the approximation of the output correlation $K_f$, the hyperparameter optimization of the kernel function is indispensable.
To estimate $K_f$, \cite{bonilla2007multi} used the EM algorithm that repeats E step where we estimate the expectation over the distribution $p(\fall | \yall, \thetav, K_f, \Sigma)$ of the missing data, i.e. $\fall$ in our case, and M step where we maximize the complete-data log-likelihood $\Lcomp \coloneq p(\yall, \fall)$, which we derive in this section.
Using the Bayes' theorem $p(\yall, \fall) = p(\yall | \fall)p(\fall)$, the complete-data log-likelihood is computed as follows:
\begin{equation}
\begin{aligned}
  \Lcomp &= \log ~\mathcal{N}(\yall | \fall, S) \mathcal{N}(\fall | \zerov_{NM}, K_{xf})~(\mathrm{Defs.}~S = I_N \otimes \Sigma, K_{xf} = K_x \otimes K_f) \\
  &= \log \frac{1}{(2\pi)^{NM} |S|^{1/2} |K_{xf}|^{1/2}}\exp\biggl(
    -\frac{(\yall - \fall)^\top S^{-1}(\yall - \fall)}{2}
    -\frac{\fall^\top K_{xf}^{-1}\fall}{2}
  \biggr) \\
  &= -NM\log 2\pi - \frac{1}{2}\log |S|
    -\frac{(\yall - \fall)^\top S^{-1}(\yall - \fall)}{2}
  - \frac{1}{2}\log |K_{xf}|
    -\frac{\fall^\top K_{xf}^{-1}\fall}{2}. \\
\end{aligned}
\end{equation}
Since $|A \otimes B| = |A|^M |B|^N$ holds for $A \in \mathbb{R}^{N \times N}$ and $B \in \mathbb{R}^{M \times M}$, we can further transform as follows:
\begin{equation}
\begin{aligned}
  \Lcomp &= - \frac{M}{2} \underbrace{\log |I_N|}_{= 0} -\frac{N}{2} \underbrace{\log |\Sigma |}_{\mathclap{=\sum_{m=1}^M \log \sigma^2_m }} -\frac{M}{2} \log |K_x| -\frac{N}{2} \log | K_f |
  -\underbrace{
    \frac{(\yall - \fall)^\top (I_N \otimes \Sigma)^{-1}(\yall - \fall)}{2}
  }_{=\frac{1}{2}\sum_{n=1}^N (\yv_n - \fv_n)^\top \Sigma^{-1} (\yv_n - \fv_n) } 
  \\
  &~~~~ - \frac{\fall^\top K_{xf}^{-1}\fall}{2}
    + \mathrm{const}. \\
    &= -\frac{N}{2} \sum_{m=1}^M \log \sigma^2_m -\frac{M}{2} \log |K_x| -\frac{N}{2} \log | K_f | -\frac{1}{2}\sum_{n=1}^N (\yv_n - \fv_n)^\top \Sigma^{-1} (\yv_n - \fv_n) - \frac{\fall^\top K_{xf}^{-1}\fall}{2} + \mathrm{const}.
\end{aligned}
\label{eq:transform-of-ll-before-last}
\end{equation}
We will finally transform the last term of \eqref{eq:transform-of-ll-before-last} using $(A \otimes B)^{-1} = A^{-1} \otimes B^{-1}$~\footnote{
  $A$ and $B$ must be a squared regular matrix.
}, \thmref{thm:otimes-and-trace}, and \thmref{thm:cyclic-property}.
Without loss of generality, we can ignore the coefficient $-1/2$ and then we can transform the last term as follows:
\begin{equation}
\begin{aligned}
  \fall^\top K_{xf}^{-1} \fall = \fall^\top (K_x \otimes K_f)^{-1} \fall &= \vectorize(F^\top)^\top (K_x^{-1} \otimes K_f^{-1}) \vectorize(F^\top) \\
  &= \trace(K_x^{-1} F K_f^{-1} F^\top)~(\because \thmref{thm:otimes-and-trace}, K_x = K_x^\top, K_f = K_f^\top) \\
  &= \trace(F^\top K_x^{-1} F K_f^{-1})~(\because \thmref{thm:cyclic-property})
\end{aligned}
\end{equation}
By plugging it back in, we obtain the complete-data log-likelihood:
\begin{equation}
\begin{aligned}
  \Lcomp = -N \sum_{m=1}^M \log \sigma_m -\frac{M}{2} \log |K_x| -\frac{N}{2} \log | K_f | - \sum_{n=1}^N \sum_{m=1}^M \frac{(y_{n,m} - f_{n,m})^2}{2\sigma_m^2}
  -\frac{1}{2} \trace(F^\top K_x^{-1} F K_f^{-1}) + \mathrm{const}.
\end{aligned}
\label{eq:complete-data-log-likelihood}
\end{equation}

\subsubsection{Maximum Likelihood Estimation in M Step}

In the M step, we maximize the complete-data log-likelihood $\Lcomp$ with respect to $\thetav$, $\Sigma$, and $K_f$.
Conventionally, we simply take hyperparameters at the stationary point.
We first estimate $\hat{\sigma}_i$:
\begin{equation}
\begin{aligned}
  \pd{\Lcomp}{\sigma_i} &= -N \sum_{m=1}^M \pd{\log \sigma_m}{\sigma_i}  - \sum_{n=1}^N \sum_{m=1}^M \pd{1/\sigma_m^2}{\sigma_i}\frac{(y_{n,m} - f_{n,m})^2}{2} 
  = -\frac{N}{\sigma_i} + \frac{1}{\sigma_i^3}\sum_{n=1}^{N} (y_{n,i} - f_{n,i})^2. 
\end{aligned}
\end{equation}
By taking the derivative of zero, we obtain:
\begin{equation}
\begin{aligned}
  \pd{\Lcomp}{\sigma_i}
  = -\frac{N}{\sigma_i} + \frac{1}{\sigma_i^3}\sum_{n=1}^{N} (y_{n,i} - f_{n,i})^2 = 0 \Longrightarrow 
  \hat{\sigma}_i^2 &= \frac{1}{N}\sum_{n=1}^{N} (y_{n,i} - f_{n,i})^2.
\end{aligned}
\end{equation}
Then we consider $K_f$.
For simplicity, we take the derivative with respect to $Q = K_f^{-1}$:
\begin{equation}
\begin{aligned}
  \pd{\Lcomp}{Q} = -\frac{N}{2} \pd{\log | Q^{-1} |}{Q} -\frac{1}{2}\pd{\trace(F^\top K_x^{-1} F Q)}{Q} = \frac{N}{2} \pd{\log |Q|}{Q} - \frac{1}{2} (F^\top K_x^{-1} F)^\top = \frac{N}{2}Q^{-1} - \frac{1}{2} F^\top K_x^{-1} F.
\end{aligned}
\label{eq:optimal-output-kernel}
\end{equation}
Note that we used the fact that $F^\top K_x^{-1} F$ and $Q^{-1}$ are symmetric for the last transformation.
By taking the derivative of zero, the following is obtained:
\begin{equation}
\begin{aligned}
  \hat{K}_f = \frac{F^\top K_x^{-1} F}{N}
\end{aligned}
\end{equation}
Finally, we consider $\thetav$.
Since we jointly optimize $\thetav$ and $K_f$, we can plug in the optimal $K_f$ in \eqref{eq:optimal-output-kernel} to \eqref{eq:complete-data-log-likelihood} and obtain the following:
\begin{equation}
\begin{aligned}
  \Lcomp &= -\frac{M}{2} \log |K_x| -\frac{N}{2} \log | K_f | 
  -\frac{1}{2} \trace(F^\top K_x^{-1} F K_f^{-1}) + \mathrm{const}. \\
  &= -\frac{M}{2} \log |K_x| -\frac{N}{2} \log~\biggl| \frac{F^\top K_x^{-1} F}{N} \biggr| 
  -\frac{1}{2} \trace(N I_M) + \mathrm{const}. \\
  &= -\frac{M}{2} \log |K_x| -\frac{N}{2} \log~|F^\top K_x^{-1} F| |N^{-M}|  + \mathrm{const}. \\
  &= -\frac{M}{2} \log |K_x| -\frac{N}{2} \log~|F^\top K_x^{-1} F|  + \mathrm{const}. \\
\end{aligned}
\end{equation}
Therefore, we need to optimize the following to yield the optimal $\thetav$:
\begin{equation}
\begin{aligned}
  \hat{\thetav} = \argmax_{\thetav} \Lcomp = \argmin_{\thetav} -2\Lcomp = \argmin_{\thetav} (M \log |K_x| + N \log~|F^\top K_x^{-1} F|).
\end{aligned}
\end{equation}
Note that this optimization must be performed numerically by using, for example, L-BFGS as analytical solutions for $\thetav$ are not available.

\subsubsection{Overview of EM Algorithm}

Although we derived the update equations for the EM algorithm in the previous section, we cannot directly obtain the solutions due to the missing data, i.e. $\fall$.
\cite{bonilla2007multi} proposed to take the expectation of each hyperparameter over $p(\fall | \yall, \thetav, K_f, \Sigma) = \mathcal{N}(\fall | K_{xf} (K_{xf} + S)^{-1}\yall,  K_{xf} - K_{xf} (K_{xf} + S)^{-1} K_{xf})$.
To summarize, we perform the following in the EM algorithm:
\begin{enumerate}
  \item Sample $\{\tilde{\fv}_{1:N}^{(k)}\}_{k=1}^K$ from $p(\fall | \yall, \hat{\thetav}^{\mathrm{old}}, \hat{K_f}^{\mathrm{old}}, \hat{\Sigma}^{\mathrm{old}})$,
  \item Optimize $\hat{\thetav}^{\mathrm{new}} = \argmin_{\thetav} (M \log |K_x| + N \log~\frac{1}{K}\sum_{k=1}^{K}|(\tilde{F}^{(k)})^\top K_x^{-1} \tilde{F}^{(k)}|)$ where $\tilde{\fv}_{1:N}^{(k)} = \vectorize((\tilde{F}^{(k)})^\top)$,
  \item Resample $\{\tilde{\fv}_{1:N}^{(k)}\}_{k=1}^K$ from $p(\fall | \yall, \hat{\thetav}^{\mathrm{new}}, \hat{K_f}^{\mathrm{old}}, \hat{\Sigma}^{\mathrm{old}})$,
  \item Calculate $(\hat{\sigma}^{\mathrm{new}}_m)^2 = \frac{1}{NK} \sum_{k=1}^{K}\sum_{n=1}^{N} (y_{n,m} - f^{(k)}_{n,m})^2$ and $\hat{K_f}^{\mathrm{new}} = \sum_{k=1}^{K}\frac{(\tilde{F}^{(k)})^\top (K_x^{\mathrm{new}})^{-1} \tilde{F}^{(k)}}{NK}$,
  \item Update the hyperparameters $\hat{\thetav}^{\mathrm{old}} \leftarrow \hat{\thetav}^{\mathrm{new}}, \hat{K_f}^{\mathrm{old}} \leftarrow \hat{K_f}^{\mathrm{new}}, \hat{\Sigma}^{\mathrm{old}} \leftarrow \hat{\Sigma}^{\mathrm{new}}$ and go back to 1.
\end{enumerate}

\subsection{Gradient Approach for Multi-Task Gaussian Process}
\label{main:method:gradient-approach}

In the previous section, we discussed the EM algorithm for MTGP, which was reported to poorly perform in terms of both the quality of solutions and the convergence speed~\citep{bonilla2007multi}.
This motivated another approach --- the approximation of $K_f$ by a gradient-based method.
Since this approach can be extended to the case where some outputs are not evaluated at some input vectors, we show the gradients for both cases, but the solutions are, in principle, identical in both cases as both of them are the maximization of the marginal log-likelihood.

\subsubsection{Case I: All Outputs Are Evaluated for Every Input Vector (Kronecker Structure)}
\label{section:gradient-kronecker}
In this case, we maximize the marginal log-likelihood $\mathcal{N}(\yall | \zerov_{NM}, K_{xf} + S)$ with respect to $\thetav$, $\Sigma$, and $K_f$.
\cite{bonilla2007multi} proposed to optimize each element in a lower-triangle matrix $L \in \mathbb{R}^{M \times M}$ so that $K_f = L L^\top$ maximizes the marginal log-likelihood:
\begin{equation}
\begin{aligned}
  L_{\mathrm{mll}} = \log \mathcal{N}(\yall | \zerov_{NM}, K_{xf} + S) &= \log \frac{1}{(2\pi)^{NM/2}|K_{xf}+S|^{1/2}}\exp\biggl(
    -\frac{\yall^\top (K_{xf}+S)^{-1} \yall}{2} 
  \biggr) \\
  &=- \frac{1}{2}\log |K_{xf}+S| -\frac{\yall^\top (K_{xf}+S)^{-1} \yall}{2} + \mathrm{const}.
\end{aligned}
\label{eq:original-marginal-log-likelihood-of-gradient-approach}
\end{equation}
We now calculate the derivative of the marginal log-likelihood with respect to a parameter $z \in \mathbb{R}$ and then this result generalizes to any other parameters such as $\thetav$, $L$, and $\Sigma$.
The derivative of the first term can be calculated as follows using $\pd{\log |A|}{z} = \trace(A^{-1}\pd{A}{z})$:
\begin{equation}
\begin{aligned}
  \pd{\log |K_{xf} + S|}{z} = \trace\biggl(
    (K_{xf} + S)^{-1}\pd{K_{xf} + S}{z}
  \biggr).
\end{aligned}
\end{equation}
The derivative of the second term can be computed using $\pd{\yv^\top A \yv}{z} = \yv^\top \pd{A}{z} \yv$ and $\pd{A^{-1}}{z} = -A^{-1}\pd{A}{z}A^{-1}$:
\begin{equation}
\begin{aligned}
  \pd{\yall^\top (K_{xf}+S)^{-1} \yall}{z} = \yall^\top \pd{(K_{xf}+S)^{-1}}{z} \yall = \yall^\top (K_{xf}+S)^{-1} \pd{K_{xf}+S}{z} (K_{xf}+S)^{-1} \yall.
\end{aligned}
\end{equation}
Therefore, if we define $G = \pd{K_{xf} + S}{z}$ and $Q = (K_{xf} + S)^{-1}$, the derivative is computed as:
\begin{equation}
\begin{aligned}
  \pd{L_{\mathrm{mll}}}{z} = -\frac{1}{2}(
    \trace(QG) + \yall^\top Q G Q \yall
  ).
\end{aligned}
\label{eq:deriv-of-mll-in-grad}
\end{equation}
Note that the derivatives of each parameter are computed as follows:
\begin{equation}
\begin{aligned}
  \pd{K_{xf} + S}{\theta_i} &= \pd{K_x}{\theta_i} \otimes I_M, \\
  \pd{K_{xf} + S}{\sigma_m^2} &= I_N \otimes \mathrm{diag}[0, \dots, \underbrace{1}_{\mathclap{m\mathrm{-th~element}}}, \dots, 0], \\
  \pd{K_{xf} + S}{L_{i,j}} &= I_N \otimes (E_{i,j}L^\top + LE_{j,i}),  \\
\end{aligned}
\label{eq:grad-of-cov-wrt-hp}
\end{equation}
where $E_{i,j} \in \mathbb{R}^{M\times M}$ is a matrix that has $0$ except at the $(i,j)$-th element where the element is $1$.
Note that the derivative with respect to the Cholesky decomposition is calculated as $\pd{K_f}{L_{i,j}} = \pd{LL^\top}{L_{i,j}} = \pd{L}{L_{i,j}}L^\top + L\pd{L^\top}{L_{i,j}} = E_{i,j}L^\top + LE_{j,i}$.
Although we listed the analytical forms of each derivative, since we can use automatic differentiation provided by PyTorch~\citep{paszke2019pytorch}, we do not have to use them explicitly.

\subsubsection{Case II: Some Outputs Are Missing for Some Input Vectors}
\label{section:gradient-non-kronecker}

This scenario is inspired by the linear coregionalization model (LCM)~\citep{wackernagel2003multivariate} and \cite{bonilla2007multi} did not discuss the extension, but we can surely extend the gradient-based approach to this setup as well.
In this setup, a set of observations on which Gaussian process is trained is different from the previous setups.
We are given a set of observations $\{\{(\xv_{n}, y_{n,m})\}_{n=1+\sum_{i=1}^{m-1}N_{i}}^{N_m}\}_{m=1}^M$ and Gaussian process is trained on this dataset where $N_m$ is the number of observations for the $m$-th output.
Simply put, some outputs may not be evaluated at some input vectors.
Even in this case, the gradient calculation is possible.
Let us define $s_m \coloneq \sum_{i=1}^{m-1}N_{i}$ where $s_1 = 0$, the $(i,j)$-th block kernel matrix as:
\begin{equation}
  K_x^{(i,j)} \coloneqq \begin{bmatrix}
    k_{\thetav}(\xv_{s_i + 1}, \xv_{s_j + 1}) & \cdots & k_{\thetav}(\xv_{s_i + 1}, \xv_{s_j + N_j}) \\
    \vdots & \ddots & \vdots \\
    k_{\thetav}(\xv_{s_i + N_i}, \xv_{s_j + 1}) & \cdots & k_{\thetav}(\xv_{s_i + N_i}, \xv_{s_j + N_j}) \\
  \end{bmatrix}
  \in \mathbb{R}^{N_i \times N_j}.
\end{equation}
Then the kernel matrix for the input vectors is defined as:
\begin{equation}
  K_x \coloneqq \begin{bmatrix}
    K_x^{(1,1)} & \dots & K_x^{(1,M)} \\
    \vdots & \ddots & \vdots \\
    K_x^{(M,1)} & \dots & K_x^{(M,M)} \\
  \end{bmatrix} \in \mathbb{R}^{s_M \times s_M}.
\end{equation}
Using Khatri-Rao product, the covariance matrix of the prior is computed as:
\begin{equation}
  K_x \ast K_f + I_{s_M} \ast \Sigma = \begin{bmatrix}
    \sigma_{1}^2 I_{N_1} + K_x^{(1,1)} (K_f)_{1,1} & \cdots & K_x^{(1,M)}(K_f)_{1,M} \\
    \vdots & \ddots & \vdots \\
    K_x^{(M,1)}(K_f)_{M,1} & \cdots & \sigma_{M}^2 I_{N_M} +  K_x^{(M,M)} (K_f)_{M,M} \\
  \end{bmatrix}
  \in \mathbb{R}^{s_M \times s_M}
  \label{eq:method:khatri-rao-product-form}
\end{equation}
where $I_{s_M}$ is partitioned in the same submatrix sizes as $K_x$, $K_x \ast K_f = (K_x^{(i,j)} \otimes (K_f)_{i,j})_{i,j}$ is Khatri-Rao product.
Note that \eqref{eq:method:khatri-rao-product-form} can also be expressed as follows:
\begin{equation}
\begin{aligned}
  K_x \ast K_f + I_{s_M} \ast \Sigma = P^\top (K_{xf} + S) P
\end{aligned}
\end{equation}
where $P \in \mathbb{R}^{NM \times s_M}$ is a matrix that is created from an identity matrix $I_N$ by removing the columns where $\yall$ does not exist.
In the NumPy style, we could write it as \texttt{P = np.eye(N * M)[:, $\sim$np.isnan(Y.flatten())]} where missing values in $Y$ are assumed to be padded by \texttt{np.nan}.
By doing so, the mathematical form of the gradient becomes very straightforward using \eqref{eq:grad-of-cov-wrt-hp} because the following holds:
\begin{equation}
\begin{aligned}
  \pd{K_x \ast K_f + I_{s_M} \ast \Sigma}{z} = \pd{P^\top (K_{xf} + S) P}{z} = P^\top\pd{K_{xf} + S}{z} P.
\end{aligned}
\end{equation}
As mentioned by \cite{lin2024scaling}, the following matrix-vector multiplication enables faster caluclation without explicitly instantiating $P$:
\begin{align}
  \mathrm{Full~Observations}:&~(A \otimes B) \vectorize (C^\top) = \underbrace{
    \vectorize (B C^\top A^\top) 
  }_{
    \hspace{-17mm}\mathrlap{
      =\texttt{(A @ C @ B.T).flatten()}      
    }
  } \\
  \mathrm{Partial~Observations}:&~P^\top (A \otimes B) P P^\top \vectorize(C^\top)
  = P^\top \vectorize (B \underbrace{
    \vectorize^{-1}(P P^\top \vectorize(C^\top))  
  }_{
    \hspace{-30mm}\mathrlap{
      =\texttt{np.where(np.isnan(C), 0.0, C).T}
    }
  }
  A^\top),
\end{align}
where $\vectorize$ and $\vectorize^{-1}$ are reshape operations, $P P^\top \vectorize(C^\top)$ is the zero padding, and $P^\top$ is the slice indexing in the NumPy coding.
Namely, the calculation is translated as \texttt{(A @ np.where(np.isnan(C), 0.0, C) @ B.T)[$\sim$np.isnan(C)]} in the NumPy style.
Although an efficient closed-form derivation for the marginal likelihood is not available anymore, the time complexity still retains $O(N^3 + M^3)$ in combination with an iterative linear system solver and the aforementioned matrix-vector multiplication.
More concretely, $(P^\top(K_{xf} + S)P)^{-1}\yv$ and $\log |P^\top(K_{xf} + S)P|$, where $\yv$ here is a vector that misses some elements in $\yall$, in \eqref{eq:original-marginal-log-likelihood-of-gradient-approach} are essential for automatic differentiation.
The former is computed by solving $\boldsymbol{\alpha} = (P^\top(K_{xf} + S)P)^{-1}\yv \Rightarrow P^\top(K_{xf} + S)P\boldsymbol{\alpha} = \yv$ with respect to $\boldsymbol{\alpha}$ by the conjugate gradient method.
The latter is computed by the Lanczos quadrature~\citep{dong2017scalable}.
The posterior sampling similarly requires $O(N^3 + M^3 + N_{\mathrm{sample}}^3)$.
Note that the efficient posterior sampling utilizes Matheron's rule~\citep{wilson2020efficiently} on top of the Kronecker structure as introduced by \cite{maddox2021bayesian}.

\section{Practical Consideration}

As mentioned by \cite{bonilla2007multi}, the gradient approach, cf. \secref{main:method:gradient-approach}, is practical in terms of both convergence speed and predictive accuracy, especially because automatic differentiation is accessible thanks to PyTorch~\citep{paszke2019pytorch} and the gradients do not have to be calculated explicitly.
Talking about the time complexity, a na\"ive implementation for optimizing the marginal log-likelihood in \eqref{eq:original-marginal-log-likelihood-of-gradient-approach} with respect to kernel hyperparameters has the time complexity of $O(N^3 M^3)$ and the memory complexity of $O(N^2 M^2)$ due to the inverse calculation of $K_{xf} + S$.
On the other hand, \cite{stegle2011efficient}, cf. Section 2.1, demonstrated that the time complexity and the memory complexity can be improved to $O(N^3 + M^3)$ and $(N^2 + M^2)$ by exploiting the Kronecker structure discussed in \secref{section:gradient-kronecker} and the diagonalizations $K_x = U_x \Lambda_x U_x^\top$ and $\Sigma^{-1/2} K_f \Sigma^{-1/2} = V_f \Lambda_f V_f^\top$~\footnote{
  Notice that we can guarantee that there exists a unitary matrix for the diagonalization of a symmetric matrix and \texttt{np.linalg.eig}, by default, returns such a unitary matrix.
}.
More specifically, using $(A \otimes B)(C \otimes D) = (AC) \otimes (BD)$ and $V_f V_f^\top = I_M$,
\begin{equation}
\begin{aligned}
  K_x \otimes K_f + I_N \otimes \Sigma &= 
  (U_x \otimes \Sigma^{1/2})
  (
    \underbrace{
      \Lambda_x \otimes \Sigma^{-1/2}K_f\Sigma^{-1/2} + I_{NM}
    }_{
      \mathclap{
        =(I_N \otimes V_f)(\Lambda_x \otimes \Lambda_f + I_{NM}) (I_N \otimes V_f^\top)
      }
    }
  )
  (U_x^\top \otimes \Sigma^{1/2}) \\
  &= (\underbrace{U_x \otimes (\Sigma^{1/2}V_f)}_{\coloneqq U_x \otimes U_f = U_{xf}})(\underbrace{\Lambda_x \otimes \Lambda_f}_{\coloneqq \Lambda_{xf}} + I_{NM}) (U_x^\top \otimes (V_f^\top\Sigma^{1/2})) \\
\end{aligned}
\end{equation}
holds, so $K_x \otimes K_f + I_N \otimes \Sigma = U_{xf} (\Lambda_{xf} + I_{NM}) U_{xf}^\top$ can be obtained.
Therefore, \eqref{eq:original-marginal-log-likelihood-of-gradient-approach} can be reformulated as follows:
\begin{equation}
\begin{aligned}
  L_{\mathrm{mll}} &=-\frac{1}{2}\log \underbrace{
    |U_{xf} (\Lambda_{xf} + I_{NM}) U_{xf}^\top|
  }_{
    \hspace{-30mm}\mathrlap{=|\Sigma|^N |\Lambda_{xf} + I_{NM}|~(\because |U_{xf}| = |U_x|^M |U_f|^N = |\Sigma|^{N/2})}
  }
  -\frac{\yall^\top (U_{xf} (\Lambda_{xf} + I_{NM}) U_{xf}^\top)^{-1} \yall}{2} + \mathrm{const} \\
  &=- \frac{1}{2}\sum_{n=1}^N \sum_{m=1}^M \log \sigma_m^2(1 + (\Lambda_x)_{n,n}(\Lambda_f)_{m,m}) -\frac{(U_{xf}^{-1} \yall)^\top (\Lambda_{xf}^{-1} + I_{NM}) U_{xf}^{-1}\yall}{2} + \mathrm{const},
\end{aligned}
\end{equation}
where $U_{xf}^{-1} = U_x^{-1} \otimes U_f^{-1} = U_x^\top \otimes (V_f^\top \Sigma^{-1/2})$.
The calculation of $U_{xf}^{-1}\yall$ can be speeded up by using $U_{xf}^{-1}\yall = \vectorize(V_f^\top\Sigma^{-1/2} Y^\top U_x)$ from $(A \otimes B)\vectorize(C^\top) = \vectorize(BC^\top A^\top)$, leading to the time complexity of $O(MN(M + N))$.
\cite{lin2024scaling} further extended this improvement to the case for non-Kronecker structure discussed in \secref{section:gradient-non-kronecker}.
The official code provided by \cite{lin2024scaling}~\footnote{
  \url{https://github.com/jandylin/Latent-Kronecker-GPs/blob/a9640af5e14610c9308508487b7270905967c0bb/models.py\#L159-L163}. (Visited on 10 July 2025)
} computes the posterior variance by direct samplings from the posterior although this is not actively discussed in the paper.
Recall that the posterior mean can be obtained via an iterative solver as explained in in \secref{section:gradient-non-kronecker}, but the official code computes both posterior mean and variance based on samples drawn from the posterior.
If additional speedup is necessary, we can use the Nystr\"om approximation~\citep{williams2000using}, which uses $K \simeq K_{\cdot, \mathcal{I}} K^{-1}_{\mathcal{I}, \mathcal{I}} K_{\mathcal{I}, \cdot}$ where $\mathcal{I} \subseteq \{1,\dots, N\}$ is a subset of indices~\footnote{
  Typically, the subset is randomly chosen.
}, and PPCA~\citep{tipping1999probabilistic}, which uses $K \simeq U\Lambda U^\top + s^2 I_N$ where $U \Lambda U^\top$ is the eigendecomposition of $K$ and $s^2$ can be analytically determined from the eigenvalues according to \cite{tipping1999probabilistic}.
Note that BoTorch~\citep{balandat2020botorch} uses $K_f \simeq \tilde{L}\tilde{L}^\top + \mathrm{diag}[v_1, \dots, v_M]$ where $\tilde{L}$ is a low-rank approximation of the Cholesky decomposition and both $\tilde{L} \in \mathbb{R}^{M \times M^\prime} (M^\prime \leq M)$ and $[v_1, \dots, v_M]$ are learned via the gradient approach~\footnote{
  By default, BoTorch uses no approximation, i.e. $M^\prime = M$.
}.

We now discuss more specific applications.
For Multi-objective optimization and constrained optimization, we can typically assume that the training dataset has the Kronecker structure, so the gradient approach from \secref{section:gradient-kronecker} can be simply applied.
For multi-fidelity optimization, although many missing data points are expected due to early stopping, many data points are still evaluated at the same input vectors.
That is why the gradient approach from \secref{section:gradient-non-kronecker} would be beneficial.
Additionally, \cite{swersky2013multi} optimized each element of the Cholesky decomposition in the log space and then exponentiated back after the optimization.
This trick benefitted them because it is reasonable to assume a positive correlation between the outputs in multi-fidelity optimization.
For meta-learning, since it is not uncommon to have no shared input vectors among each task, it would be helpful to use approximation methods such as the Nystr\"om approximation.

With that being said, if many evaluations are expected during an optimization, we should consider using a light-weight BO method such as TPE extensions instead of GPBO such as MOTPE~\citep{ozaki2020multiobjective,ozaki2022multiobjective} for multi-objective optimization, \citep{watanabe2022c,watanabe2023c} for constrained optimization, \citep{falkner2018bohb} for multi-fidelity optimization, \citep{watanabe2022multi,watanabe2023speeding} for meta-learning.

\section{Conclusion}
In this paper, we showed the detailed derivations of the intermediate processes for MTGP.
We hope that our derivations help practitioners verify their codes and will be a helpful reference for researchers in the future.

\newpage
\bibliographystyle{apalike}  
\bibliography{ref}

\newpage
\appendix

\section{Preliminaries}

\subsection{Basic Theorems in Linear Algebra}

In this section, we consistently assume that $A \in \mathbb{R}^{N \times N}$, $B \in \mathbb{R}^{M \times M}$, $C \in \mathbb{R}^{N \times M}$, and $D \in \mathbb{R}^{M \times N}$.

\begin{theorem}
  $\vectorize(C^\top)^\top (A \otimes B) \vectorize(C^\top) = \trace(ACB^\top C^\top)$ holds.
  \label{thm:otimes-and-trace}
\end{theorem}
The proof is available in \appxref{proof:otimes-and-trace}.

\begin{theorem}[Cyclic Property]
  $\trace(CD) = \trace(DC)$ holds.
  \label{thm:cyclic-property}
\end{theorem}
The proof is available in \appxref{proof:cyclic-property}.
Furthermore, the following facts will be used in this paper although we omit concrete proofs:
\begin{itemize}
  \vspace{-1mm}
  \item $|A \otimes B| = |A|^M |B|^N$,
  \vspace{-1mm}
  \item $(A \otimes B)^{-1} = A^{-1} \otimes B^{-1}$,
  \vspace{-1mm}
  \item $(A \otimes B)(C \otimes D) = (AC) \otimes (BD)$,
  \vspace{-1mm}
  \item $(A \otimes B)\vectorize(C^\top) = \vectorize(BC^\top A^\top)$,
  \vspace{-1mm}
  \item $\pd{\log |A|}{z} = \trace(A^{-1}\pd{A}{z})$,
  \vspace{-1mm}
  \item $\pd{\yv^\top A \yv}{z} = \yv^\top \pd{A}{z} \yv$, and
  \vspace{-1mm}
  \item $\pd{A^{-1}}{z} = -A^{-1}\pd{A}{z}A^{-1}$.
  \vspace{-1mm}
\end{itemize}

\subsection{Gaussian Process}
In the simplest setup, GP estimates the posterior distribution of a single-output function $f$ at a point $\xnew$.
Since the posterior distribution is assumed to be a Gaussian distribution, the posterior mean and variance are of interest and they will be calculated as follows:
\begin{equation}
\begin{aligned}
  \mathcal{N}(k_{\mathrm{new}}^\top(K_x + \sigma^2 I)^{-1} \yv, c_{\mathrm{new}} -k_{\mathrm{new}}^\top (K_x + \sigma^2 I)^{-1} k_{\mathrm{new}}),
\end{aligned}
\end{equation}
where $k_{\mathrm{new}} \coloneqq [k_{\thetav}(\xv_1, \xnew), \dots, k_{\thetav}(\xv_N, \xnew)]$ and $c_{\mathrm{new}} \coloneqq k_{\thetav}(\xnew, \xnew)$.
The kernel hyperparameters $\thetav$ are mostly optimized by maximizing the following marginal log-likelihood (typically plus a log hyperparameter prior):
\begin{equation}
\begin{aligned}
  \log p(\yv | \thetav) = -\frac{1}{2} \log |K_x + \sigma^2 I| -\frac{1}{2} \yv^\top (K_x + \sigma^2 I)^{-1} \yv + \mathrm{const}.
\end{aligned}
\end{equation}
The optimization is practically performed by the gradient approach.
This paper discusses the kernel hyperparameter optimization for a multi-output function.

\section{Proofs}
\subsection{Proof of \thmref{thm:otimes-and-trace}}
\label{proof:otimes-and-trace}

\begin{proof}
  We first calculate the right hand side.
  The matrix multiplication can be easily computed as follows:
  \begin{equation}
    \begin{aligned}
    (AC)_{i,j} &= \sum_{k=1}^{N} A_{i,k}C_{k,j} \in \mathbb{R}^{N \times M},
    (B^\top C^\top)_{i,j} &= \sum_{k = 1}^{M} B_{k, i} C_{j, k} \in \mathbb{R}^{M \times N}. \\
  \end{aligned}
  \end{equation}
  Therefore, we yield:
  \begin{equation}
  \begin{aligned}
    (ACB^\top C^\top)_{i,j} &= \sum_{k=1}^{M} (AC)_{i,k} (B^\top C^\top)_{k,j} = \sum_{p=1}^{M} \sum_{q=1}^{N} \sum_{r = 1}^{M} A_{i,q}C_{q,p} B_{r, p} C_{j, r}, \\
    \mathrm{RHS} &= \trace(ACB^\top C^\top) = \sum_{k=1}^{N} (ACB^\top C^\top)_{k,k} = \sum_{k=1}^{N} \sum_{p=1}^{M} \sum_{q=1}^{N} \sum_{r = 1}^{M} A_{k,q}C_{q,p} B_{r, p} C_{k, r}.
  \end{aligned}
  \end{equation}
  We now consider the $(i,j)$-th block of $A \otimes B$ in the left hand side, i.e. $A_{i,j}B [C_{j,1}, \dots, C_{j,M}]^\top$.
  Then we obtain $\sum_{p=1}^M A_{i,j}B_{r,p}C_{j,p}$ at the $r$-th row.
  Since we are taking the inner product, we need to sum up the $r$-th row elements in the $(i,j)$-th blocks for $j \in \{1,\dots,N\}$, leading to $\sum_{j=1}^{N}\sum_{p=1}^M A_{i,j}B_{r,p}C_{j,p}$.
  By considering the summation over the rows $(r \in \{1,\dots,M\})$ in the $i$-th blocks of $\sum_{j=1}^{N}\sum_{p=1}^M A_{i,j}B_{r,p}C_{j,p}$ and $\vectorize(C^\top)$, we yield $\sum_{j=1}^{N}\sum_{p=1}^M \sum_{r=1}^{M} C_{i,r} A_{i,j}B_{r,p}C_{j,p}$ at the $i$-th block.
  By adding up from $i = 1$ to $N$, we obtain:
  \begin{equation}
  \begin{aligned}
    \mathrm{RHS} &= \sum_{i=1}^{N}\sum_{j=1}^{N}\sum_{p=1}^M \sum_{r=1}^{M} C_{i,r} A_{i,j}B_{r,p}C_{j,p} \\
    &= \sum_{k=1}^{N}\sum_{q=1}^{N}\sum_{p=1}^M \sum_{r=1}^{M} A_{k,q} C_{q,p} B_{r,p} C_{k,r} = \mathrm{LHS}.
  \end{aligned}
  \end{equation}
  This completes the proof.
\end{proof}

\subsection{Proof of \thmref{thm:cyclic-property}}
\label{proof:cyclic-property}

\begin{proof}
  We calculate both hand sides, respectively:
  \begin{equation}
  \begin{aligned}
    \mathrm{LHS} = \sum_{n=1}^{N} \sum_{m=1}^M C_{n,m}D_{m,n},
    \mathrm{RHS} = \sum_{m=1}^{M} \sum_{n=1}^N D_{m,n}C_{n,m}
  \end{aligned}
  \end{equation}
  Since the results of the left and right hand sides match, this completes the proof.
\end{proof}


\end{document}

%% file: settings/preamble.tex
\input{settings/packages.tex}
\input{settings/layout.tex}  
\input{settings/paper-status.tex}

\input{settings/user-commands.tex}

%% file: settings/packages.tex
\usepackage[final]{automl}
\usepackage{microtype} 
\usepackage{booktabs}  
\usepackage{url}  
\usepackage{csquotes}
\usepackage{natbib}

\usepackage[utf8]{inputenc}
\usepackage{url}

\usepackage{amsmath}
\usepackage{bbm}
\usepackage{bm}
\usepackage{cancel}
\usepackage{mathtools}

\usepackage{ascmac}  
\usepackage{algorithm}
\usepackage{algpseudocode}

\usepackage{fancyhdr}
\usepackage{tcolorbox}
\usepackage{tikz}

\usepackage{booktabs}
\usepackage{comment}
\usepackage{lastpage}
\usepackage{listings}
\usepackage{multibib}
\usepackage{multirow}
\usepackage[super]{nth}

\tcbuselibrary{breakable, skins, theorems}

\allowdisplaybreaks

\newtheorem{theorem}{Theorem}

\newtheorem{proof}{Proof}

\algtext*{EndFor}
\algtext*{EndWhile}
\algtext*{EndIf}
\algtext*{EndProcedure}
\algtext*{EndFunction}
\algnewcommand{\LineComment}[1]{\State \(\triangleright\) #1}

\newcites{appx}{References}

\hypersetup{%
  pdfauthor={Shuhei Watanabe}, 
  pdftitle={Derivation of Multi-Task GP},
  pdfsubject={},
  pdfkeywords={AutoML, LaTeX, style}
}

%% file: settings/paper-status.tex
\newif\ifunderreview
\newif\ifsubmission
\newif\ifappendix

\underreviewfalse

\submissiontrue

\appendixtrue

\ifsubmission
\newcommand{\todo}[1]{}
\newcommand{\replace}[2]{}
\newcommand{\sw}[1]{}

\else
\newcommand{\todo}[1]{\textbf{\textcolor{red}{[TODO: #1]}}}

\newcommand{\replace}[2]{\textbf{\textcolor{red}{[del: \cancel{#1}]}}\textbf{\textcolor{blue}{[new: #2]}}}

\newcommand{\sw}[1]{\textbf{\textcolor{cyan}{[SW: #1]}}}

\usepackage[inline]{showlabels}

\fi

%% file: settings/user-commands.tex
\makeatletter
\newcommand{\customlabel}[2]{%
   \protected@write \@auxout {}{\string \newlabel {#1}{{#2}{\thepage}{#2}{#1}{}} }%
   \hypertarget{#1}{}
}
\makeatother

\newcommand{\argmin}{\mathop{\mathrm{argmin}}}
\newcommand{\argmax}{\mathop{\mathrm{argmax}}}

\newcommand{\secref}[1]{Section~\ref{#1}}
\renewcommand{\eqref}[1]{Eq.~(\ref{#1})}

\DeclareFixedFont{\ttb}{T1}{txtt}{bx}{n}{12} 
\DeclareFixedFont{\ttm}{T1}{txtt}{m}{n}{12}  

\definecolor{deepblue}{rgb}{0,0,0.5}
\definecolor{deepred}{rgb}{0.6,0,0}
\definecolor{deepgreen}{rgb}{0,0.5,0}

%% file: settings/author-info.tex
\title{
  Derivation of Output Correlation Inferences for \\
  Multi-Output (aka Multi-Task) Gaussian Process
}

\author{
  Shuhei Watanabe \\
  Preferred Networks Inc. \\
  \texttt{shuheiwatanabe@preferred.jp}
}

\maketitle

%% file: sections/abstract/main.tex
\begin{abstract}
  Gaussian process (GP) is arguably one of the most widely used machine learning algorithms in practice.
  One of its prominent applications is Bayesian optimization (BO).
  Although the vanilla GP itself is already a powerful tool for BO, it is often beneficial to be able to consider the dependencies of multiple outputs.
  To do so, Multi-task GP (MTGP) is formulated, but it is not trivial to fully understand the derivations of its formulations and their gradients from the previous literature.
  This paper serves friendly derivations of the MTGP formulations and their gradients.
\end{abstract}

%% file: submission.bbl
\begin{thebibliography}{}

\bibitem[Alvarez and Lawrence, 2008]{alvarez2008sparse}
Alvarez, M. and Lawrence, N. (2008).
\newblock Sparse convolved {G}aussian processes for multi-output regression.
\newblock {\em Advances in Neural Information Processing Systems}.

\bibitem[Balandat et~al., 2020]{balandat2020botorch}
Balandat, M., Karrer, B., Jiang, D., Daulton, S., Letham, B., Wilson, A., and
  Bakshy, E. (2020).
\newblock {BoTorch: A Framework for Efficient Monte-Carlo Bayesian
  Optimization}.
\newblock In {\em Advances in Neural Information Processing Systems}.

\bibitem[Bergstra et~al., 2011]{bergstra2011algorithms}
Bergstra, J., Bardenet, R., Bengio, Y., and K{\'e}gl, B. (2011).
\newblock Algorithms for hyper-parameter optimization.
\newblock {\em Advances in Neural Information Processing Systems}.

\bibitem[Bonilla et~al., 2007]{bonilla2007multi}
Bonilla, E., Chai, K., and Williams, C. (2007).
\newblock Multi-task {G}aussian process prediction.
\newblock {\em Advances in neural information processing systems}.

\bibitem[Brochu et~al., 2010]{brochu2010tutorial}
Brochu, E., Cora, V., and de~Freitas, N. (2010).
\newblock A tutorial on {Bayesian} optimization of expensive cost functions,
  with application to active user modeling and hierarchical reinforcement
  learning.
\newblock {\em arXiv:1012.2599}.

\bibitem[Daulton et~al., 2020]{daulton2020differentiable}
Daulton, S., Balandat, M., and Bakshy, E. (2020).
\newblock Differentiable expected hypervolume improvement for parallel
  multi-objective {B}ayesian optimization.
\newblock {\em Advances in Neural Information Processing Systems}.

\bibitem[Daulton et~al., 2022]{daulton2022multi}
Daulton, S., Eriksson, D., Balandat, M., and Bakshy, E. (2022).
\newblock Multi-objective {B}ayesian optimization over high-dimensional search
  spaces.
\newblock In {\em Uncertainty in Artificial Intelligence}.

\bibitem[Dong et~al., 2017]{dong2017scalable}
Dong, K., Eriksson, D., Nickisch, H., Bindel, D., and Wilson, A. (2017).
\newblock Scalable log determinants for {G}aussian process kernel learning.
\newblock {\em Advances in Neural Information Processing Systems}.

\bibitem[Eriksson and Poloczek, 2021]{eriksson2021scalable}
Eriksson, D. and Poloczek, M. (2021).
\newblock Scalable constrained {Bayesian} optimization.
\newblock In {\em International Conference on Artificial Intelligence and
  Statistics}.

\bibitem[Falkner et~al., 2018]{falkner2018bohb}
Falkner, S., Klein, A., and Hutter, F. (2018).
\newblock {BOHB}: Robust and efficient hyperparameter optimization at scale.
\newblock In {\em International Conference on Machine Learning}.

\bibitem[Feurer et~al., 2018]{feurer2018practical}
Feurer, M., Letham, B., Hutter, F., and Bakshy, E. (2018).
\newblock Practical transfer learning for {Bayesian} optimization.
\newblock {\em arXiv:1802.02219}.

\bibitem[Gardner et~al., 2014]{gardner2014bayesian}
Gardner, J., Kusner, M., Xu, Z., Weinberger, K., and Cunningham, J. (2014).
\newblock Bayesian optimization with inequality constraints.
\newblock In {\em International Conference on Machine Learning}.

\bibitem[Garnett, 2022]{garnett2022bayesian}
Garnett, R. (2022).
\newblock {\em {Bayesian Optimization}}.
\newblock Cambridge University Press.

\bibitem[Gelbart et~al., 2014]{gelbart2014bayesian}
Gelbart, M., Snoek, J., and Adams, R. (2014).
\newblock Bayesian optimization with unknown constraints.
\newblock {\em arXiv:1403.5607}.

\bibitem[Kandasamy et~al., 2019]{kandasamy2019multi}
Kandasamy, K., Dasarathy, G., Oliva, J., Schneider, J., and Poczos, B. (2019).
\newblock Multi-fidelity {G}aussian process bandit optimisation.
\newblock {\em Journal of Artificial Intelligence Research}, 66.

\bibitem[Kandasamy et~al., 2017]{kandasamy2017multi}
Kandasamy, K., Dasarathy, G., Schneider, J., and P{\'o}czos, B. (2017).
\newblock Multi-fidelity {B}ayesian optimisation with continuous
  approximations.
\newblock In {\em International Conference on Machine Learning}.

\bibitem[Lin et~al., 2024]{lin2024scaling}
Lin, J., Ament, S., Balandat, M., and Bakshy, E. (2024).
\newblock Scaling {G}aussian processes for learning curve prediction via latent
  {K}ronecker structure.
\newblock {\em arXiv:2410.09239}.

\bibitem[Lindauer et~al., 2022]{lindauer2022smac3}
Lindauer, M., Eggensperger, K., Feurer, M., Biedenkapp, A., Deng, D.,
  Benjamins, C., Ruhkopf, T., Sass, R., and Hutter, F. (2022).
\newblock {SMAC3}: A versatile {B}ayesian optimization package for
  hyperparameter optimization.
\newblock {\em Journal of Machine Learning Research}, 23.

\bibitem[Maddox et~al., 2021]{maddox2021bayesian}
Maddox, W., Balandat, M., Wilson, A., and Bakshy, E. (2021).
\newblock {B}ayesian optimization with high-dimensional outputs.
\newblock {\em Advances in neural information processing systems}.

\bibitem[McLachlan and Krishnan, 2008]{mclachlan2008algorithm}
McLachlan, J. and Krishnan, T. (2008).
\newblock {\em The {EM} algorithm and extensions}.
\newblock John Wiley \& Sons.

\bibitem[Nguyen et~al., 2014]{nguyen2014collaborative}
Nguyen, T., Bonilla, E., et~al. (2014).
\newblock Collaborative multi-output {G}aussian processes.
\newblock In {\em Uncertainty in Artificial Intelligence}.

\bibitem[Ozaki et~al., 2022]{ozaki2022multiobjective}
Ozaki, Y., Tanigaki, Y., Watanabe, S., Nomura, M., and Onishi, M. (2022).
\newblock Multiobjective tree-structured {P}arzen estimator.
\newblock {\em Journal of Artificial Intelligence Research}, 73.

\bibitem[Ozaki et~al., 2020]{ozaki2020multiobjective}
Ozaki, Y., Tanigaki, Y., Watanabe, S., and Onishi, M. (2020).
\newblock Multiobjective tree-structured {Parzen} estimator for computationally
  expensive optimization problems.
\newblock In {\em Genetic and Evolutionary Computation Conference}.

\bibitem[Paszke et~al., 2019]{paszke2019pytorch}
Paszke, A., Gross, S., Massa, F., Lerer, A., Bradbury, J., Chanan, G., Killeen,
  T., Lin, Z., Gimelshein, N., Antiga, L., et~al. (2019).
\newblock {PyTorch}: An imperative style, high-performance deep learning
  library.
\newblock {\em Advances in neural information processing systems}.

\bibitem[Shahriari et~al., 2016]{shahriari2016taking}
Shahriari, B., Swersky, K., Wang, Z., Adams, R., and de~Freitas, N. (2016).
\newblock Taking the human out of the loop: {A} review of {B}ayesian
  optimization.
\newblock {\em Proceedings of the {IEEE}}, 104.

\bibitem[Song et~al., 2019]{song2019general}
Song, J., Chen, Y., and Yue, Y. (2019).
\newblock A general framework for multi-fidelity {B}ayesian optimization with
  {G}aussian processes.
\newblock In {\em International Conference on Artificial Intelligence and
  Statistics}.

\bibitem[Stegle et~al., 2011]{stegle2011efficient}
Stegle, O., Lippert, C., Mooij, J., Lawrence, N., and Borgwardt, K. (2011).
\newblock Efficient inference in matrix-variate {G}aussian models with iid
  observation noise.
\newblock {\em Advances in Neural Information Processing Systems}.

\bibitem[Swersky et~al., 2013]{swersky2013multi}
Swersky, K., Snoek, J., and Adams, R. (2013).
\newblock Multi-task {Bayesian} optimization.
\newblock In {\em Advances in Neural Information Processing Systems}.

\bibitem[Tipping and Bishop, 1999]{tipping1999probabilistic}
Tipping, M. and Bishop, C. (1999).
\newblock Probabilistic principal component analysis.
\newblock {\em Journal of the Royal Statistical Society Series B: Statistical
  Methodology}, 61.

\bibitem[Titsias, 2009]{titsias2009variational}
Titsias, M. (2009).
\newblock Variational learning of inducing variables in sparse {G}aussian
  processes.
\newblock In {\em Artificial Intelligence and Statistics}.

\bibitem[Wackernagel, 2003]{wackernagel2003multivariate}
Wackernagel, H. (2003).
\newblock {\em Multivariate geostatistics: an introduction with applications}.
\newblock Springer Science \& Business Media.

\bibitem[Watanabe, 2023]{watanabe2023tree}
Watanabe, S. (2023).
\newblock Tree-structured {P}arzen estimator: Understanding its algorithm
  components and their roles for better empirical performance.
\newblock {\em arXiv:2304.11127}.

\bibitem[Watanabe et~al., 2022]{watanabe2022multi}
Watanabe, S., Awad, N., Onishi, M., and Hutter, F. (2022).
\newblock Multi-objective tree-structured {P}arzen estimator meets
  meta-learning.
\newblock {\em arXiv:2212.06751}.

\bibitem[Watanabe et~al., 2023]{watanabe2023speeding}
Watanabe, S., Awad, N., Onishi, M., and Hutter, F. (2023).
\newblock Speeding up multi-objective hyperparameter optimization by task
  similarity-based meta-learning for the tree-structured {P}arzen estimator.
\newblock {\em arXiv:2212.06751}.

\bibitem[Watanabe and Hutter, 2022]{watanabe2022c}
Watanabe, S. and Hutter, F. (2022).
\newblock c-{TPE}: Generalizing tree-structured {P}arzen estimator with
  inequality constraints for continuous and categorical hyperparameter
  optimization.
\newblock {\em arXiv:2211.14411}.

\bibitem[Watanabe and Hutter, 2023]{watanabe2023c}
Watanabe, S. and Hutter, F. (2023).
\newblock c-{TPE}: Tree-structured {P}arzen estimator with inequality
  constraints for expensive hyperparameter optimization.
\newblock {\em arXiv:2211.14411}.

\bibitem[Williams and Seeger, 2000]{williams2000using}
Williams, C. and Seeger, M. (2000).
\newblock Using the {N}ystr{\"o}m method to speed up kernel machines.
\newblock {\em Advances in Neural Information Processing Systems}.

\bibitem[Wilson et~al., 2011]{wilson2011gaussian}
Wilson, A., Knowles, D., and Ghahramani, Z. (2011).
\newblock {G}aussian process regression networks.
\newblock {\em arXiv:1110.4411}.

\bibitem[Wilson et~al., 2020]{wilson2020efficiently}
Wilson, J., Borovitskiy, V., Terenin, A., Mostowsky, P., and Deisenroth, M.
  (2020).
\newblock Efficiently sampling functions from {G}aussian process posteriors.
\newblock In {\em International Conference on Machine Learning}.

\bibitem[Wistuba et~al., 2022]{wistuba2022supervising}
Wistuba, M., Kadra, A., and Grabocka, J. (2022).
\newblock Supervising the multi-fidelity race of hyperparameter configurations.
\newblock {\em Advances in Neural Information Processing Systems}.

\bibitem[Yang et~al., 2019]{yang2019multi}
Yang, K., Emmerich, M., Deutz, A., and B{\"a}ck, T. (2019).
\newblock Multi-objective {B}ayesian global optimization using expected
  hypervolume improvement gradient.
\newblock {\em Swarm and evolutionary computation}, 44.

\end{thebibliography}
